\documentclass[conference]{IEEEtran}
\IEEEoverridecommandlockouts

\usepackage{amsmath,amssymb,amsfonts}
\usepackage{algorithmic}
\usepackage{array}
\usepackage{graphicx}
\usepackage{textcomp}
\usepackage{xcolor}
\usepackage{bm}
\usepackage{bbm}
\usepackage{multirow}
\usepackage{booktabs}
\usepackage{algorithm}
\usepackage{microtype}
\usepackage{eqparbox}
\usepackage{url}

\usepackage{subcaption}

\usepackage{relsize}
\usepackage{hyperref}

\usepackage[utf8]{inputenc}
\DeclareUnicodeCharacter{0308}{}
\DeclareUnicodeCharacter{0301}{}

\usepackage[backend=biber,style=ieee,sorting=none, maxbibnames=10]{biblatex}
\usepackage{microtype}

\renewcommand{\algorithmiccomment}[1]{\bgroup\hfill$\triangleright$~#1\egroup}
\newcommand{\algorithmiccommentn}[1]{\bgroup$\triangleright$~#1\egroup}

\newcommand{\argmax}[1]{\underset{#1}{\operatorname{arg}\,\operatorname{max}}\;}
\DeclareMathOperator*{\argmaxil}{argmax}

\begin{document}
	
	

\title{Learning on a Budget via Teacher Imitation}

\author{\IEEEauthorblockN{Erc\"{u}ment \.{I}lhan, Jeremy Gow and Diego Perez-Liebana}
\IEEEauthorblockA{\textit{School of Electronic Engineering and Computer Science} \\
\textit{Queen Mary University of London}\\
London, United Kingdom \\
\{e.ilhan, jeremy.gow, diego.perez\}@qmul.ac.uk}
}

\IEEEpubid{\begin{minipage}{\textwidth}\ \\[12pt]
978-1-6654-3886-5/21/\$31.00 \copyright 2021 IEEE
\end{minipage}}

\maketitle

\begin{abstract}
Deep Reinforcement Learning (RL) techniques can benefit greatly from leveraging prior experience, which can be either self-generated or acquired from other entities.
Action advising is a framework that provides a flexible way to transfer such knowledge in the form of actions between teacher-student peers.
However, due to the realistic concerns, the number of these interactions is limited with a budget; therefore, it is crucial to perform these in the most appropriate moments.
There have been several promising studies recently that address this problem setting especially from the student's perspective.
Despite their success, they have some shortcomings when it comes to the practical applicability and integrity as an overall solution to the learning from advice challenge.
In this paper, we extend the idea of advice reusing via teacher imitation to construct a unified approach that addresses both advice collection and advice utilisation problems.
We also propose a method to automatically tune the relevant hyperparameters of these components on-the-fly to make it able to adapt to any task with minimal human intervention.
The experiments we performed in $5$ different Atari games verify that our algorithm either surpasses or performs on-par with its top competitors while being far simpler to be employed.
Furthermore, its individual components are also found to be providing significant advantages alone.
\end{abstract}


\section{Introduction} \label{sec:introduction}

Deep Reinforcement Learning (RL) has been proven to be a successful approach to solve decision-making problems in a variety of difficult domains such as video games \cite{DBLP:journals/nature/VinyalsBCMDCCPE19}, board games \cite{DBLP:journals/corr/abs-1911-08265} or robot manipulation \cite{DBLP:journals/corr/abs-1910-07113}.
However, achieving the reported performances is not entirely straightforward.
One of the most critical setbacks in deep RL is the exhaustive training processes that usually require many interactions with the environment.
This occurs mainly due to the RL inherent exploration challenges as well as the complexity of the incorporated function approximators, e.g. deep neural networks.
To this date, there has been a remarkable amount of research effort to overcome the sample inefficiency by devising advanced exploration strategies \cite{DBLP:journals/corr/abs-1908-02388}.
In addition to these, other lines of work that focus on leveraging some legacy knowledge to tackle these issues have also been studied extensively with great success.

The ability to learn by utilising the prior experience instead of starting from scratch is an essential component of intelligence.
In RL, this idea has been investigated in various forms that are tailored for different problem settings.
Imitation Learning (IL) \cite{DBLP:journals/neco/Pomerleau91} studies the concept of mimicking an expert behaviour presented in a pre-recorded dataset without allowing any RL rewards from the environment itself.
Similarly, Learning from Demonstrations (LfD) \cite{DBLP:conf/nips/Schaal96}\cite{DBLP:conf/aaai/HesterVPLSPHQSO18} extends this idea to incorporate the RL interactions and rewards to further surpass the experts.
Some other approaches such as Policy Reuse \cite{DBLP:conf/atal/FernandezV06} study the ways of speeding up the agent's learning by leveraging the past policies directly, instead of datasets.

In this paper, we study a different problem setup where it is not possible to have any pre-generated datasets or to directly obtain the useful policies themselves; instead, the learning agent only has access to some peer(s) over a limited communication channel.
Such a setting is especially relevant for the scenarios with unknown task specifications (prior to learning and deployment) and non-transferable yet beneficial policies, e.g. humans, non-stationary agents, in the loop.
A flexible framework that is tailored for this setting is called Action Advising \cite{DBLP:conf/atal/TorreyT13}.
According to this, agents exchange knowledge between each other in the form of actions to speed-up their learning progressions.
However, the number of these peer-to-peer interactions are limited with a budget to resemble the real-world limitations, which essentially converts the problem into determining the best possible way to utilise the available budget.
Based on the peer that is in charge of driving these interactions, action advising algorithms can either be student-initiated, teacher-initiated or jointly-initiated.

Action advising algorithms in deep RL have obtained promising results with an emphasis on student-initiated strategies \cite{DBLP:conf/cig/IlhanGP19}\cite{DBLP:conf/aaai/SilvaHKT20a}\cite{ilhan2020studentinitiated}.
While the majority of these focus on addressing \emph{when to ask for advice} question, some recent approaches \cite{aawithimitation} have also investigated the ways to further utilise the collected advice by imitating and reusing the teacher policy.
Despite these developments, there are currently several significant shortcomings present.
These techniques often employ some threshold hyperparameters to control the decisions to initiate advice exchange interactions which play a key role in their efficiencies.
However, these parameters are sensitive to the learning state of the models as well as the domain properties.
Therefore, they need to be tuned very carefully prior to execution, which involve unrealistically accessing the target tasks for trial runs.
Furthermore, the studies to further leverage the teacher advice beyond collection are currently in their early stages and do not provide a complete solution to the problem besides addressing the advice reusing aspect.

In this paper, we present an all-in-one student-initiated approach that is capable of collecting and reusing advice in a budget-efficient manner, by extending \cite{aawithimitation} in multiple ways.
First, we propose a method for automatically determining the threshold parameters responsible for the decisions to request and reuse advice.
This greatly alleviates the burden of task-specific hyperparameter tuning procedures.
Secondly, we follow a decaying advice reuse schedule that is not tied to the student's exploration strategy.
Finally, instead of using the imitated policy only for reusing advice as in \cite{aawithimitation}, we incorporate this policy to determine and collect more diverse advice to construct a more universal imitation policy.

The rest of this paper is structured as follows:
Section~\ref{sec:related_work} outlines the most relevant previous work.
In Section~\ref{sec:background}, the background knowledge that is required to understand the paper is provided.
Afterwards, we describe our approach in detail in Section~\ref{sec:proposed_algorithm}.
Then, Section~\ref{sec:experiments} presents our evaluation domain and experimental procedure.
In Section~\ref{sec:results}, we share and analyse the experiment results.
Finally, Section~\ref{sec:conclusions} wraps up this study with conclusions and future remarks.

\section{Related Work} \label{sec:related_work}

Action advising techniques with budget constraints were originally invented and studied extensively in tabular domains.
In \cite{DBLP:conf/atal/TorreyT13}, the teacher-student learning procedure was formalised for the first time and some solutions from the teacher's perspective were proposed.
This was later extended with some new heuristics \cite{DBLP:journals/connection/TaylorCFVT14} as well as with a meta-learning approach \cite{zimmer2014teacher}.
Later on, the action advising interactions are also studied in student-initiated and jointly-initiated forms \cite{DBLP:conf/ijcai/AmirKKG16} which were also adopted in multi-agent problems where the agents take student and teacher roles interchangeably \cite{DBLP:conf/atal/SilvaGC17}.
In a recent work \cite{DBLP:conf/atal/ZhuCLH20}, the idea of reusing the previously collected advice is investigated to improve the learning performance and also to make a more efficient use of the available budget.

Deep RL is a considerably new domain for the action advising studies.
\cite{DBLP:journals/corr/abs-1812-02632} introduced a novel LfD setup in which the demonstrations dataset is built interactively as in action advising.
To do so, they employed uncertainty estimation capable models with LfD loss terms integrated in the learning stage.
\cite{DBLP:conf/cig/IlhanGP19} extended the jointly-initiated action advising \cite{DBLP:conf/atal/SilvaGC17} to be applicable in multi-agent deep RL for the first time.
This study replaced the state counters that were used to assess uncertainty in the tabular version \cite{DBLP:conf/atal/SilvaGC17} with state novelty measurements with the aid of Random Network Distillation (RND) \cite{DBLP:journals/corr/abs-1810-12894}.
In \cite{DBLP:conf/aaai/SilvaHKT20a}, an uncertainty-based advice collection strategy was proposed.
According to this, the student adopts a multi-headed neural network architecture to access epistemic uncertainty estimations as in \cite{DBLP:journals/corr/abs-1812-02632}.
Later, \cite{ilhan2020studentinitiated} further studied the state novelty-based idea of \cite{DBLP:conf/cig/IlhanGP19} to devise a better student-initiated advice collection method.
Specifically, they made RND module to be updated only for the states that are involved in advice collection.
That way, the student was ensured to benefit from the teacher regardless of its own knowledge about the states to tackle the special cases of belated inclusion of the teacher.
More recently, \cite{aawithimitation} studied the advice reuse idea in deep RL.
In this work, an imitation model of the teacher is constructed partially with the collected advice.
Furthermore, Dropout regularisation \cite{DBLP:conf/icml/GalG16} is also incorporated in this model to enable it to make uncertainty-aware decisions when it comes to reusing the self-generated advice.

Among these studies in deep RL, only \cite{DBLP:journals/corr/abs-1812-02632} and \cite{aawithimitation} investigated the concept of advice reuse.
Our paper differs from them in several ways.
The idea in \cite{DBLP:journals/corr/abs-1812-02632} require using uncertainty estimation capable models in the student's RL algorithm.
Moreover, the RL algorithm also needs to be modified to have the LfD loss terms.
In contrast, the algorithm we propose does not require the student to have any specific RL models or loss functions.
Thus, the student agent can be treated as a blackbox, which lets our method to be applied to a wider range of agent types.
\cite{aawithimitation} performs advice reuse as a separate module, as we described here.
However, they rely on the previously proposed advice collection strategies instead of taking advantage of its own imitation module to manage the advice collection process via uncertainty.
Furthermore, some of the hyperparameters in \cite{aawithimitation} limit its practical applicability due to being difficult to tune, which we also address in this work.

\section{Background} \label{sec:background}

\subsection{Reinforcement Learning and Deep Q-Networks}

Reinforcement Learning (RL) \cite{sutton2018reinforcement} is a trial-and-error learning paradigm that studies decision-making problems where an agent learns to accomplish a task.
This interaction within the environment is commonly formalised as a Markov Decision Process (MDP).
In MDP, the environment is defined with the tuple $\langle \mathcal{S}, \mathcal{A}, R, \mathcal{T}, \gamma \rangle$ where $\mathcal{S}$ is the set of states, $\mathcal{A}$ is the set of available actions, $R(s, a)$ is the reward function, $\mathcal{T}(s \vert a, s')$ defines the transition probabilities, and $\gamma \in [0,1]$ is the discount factor.
At every timestep $t$, the agent applies an action $a_t$ in state $s_t$ to advance to the next state $s_{t+1}$ while receiving a reward $r_t$.
These actions are determined by the agent's policy $\pi \colon \mathcal{S} \rightarrow \mathcal{A}$, and RL's goal is to learn an agent policy $\pi_{\theta}$ (with parameters $\theta$) that maximises the cumulative discounted sum of the rewards $\sum_{t=0}^{\infty} \gamma^{t} r_{t}$.
There are various different approaches in the RL literature to achieve this.
For instance, the well-known Q-learning algorithm does so by learning the state-action values $Q(s, a)$ via Bellman equations \cite{sutton2018reinforcement} and making the agent follow the policy $\pi(s) = \argmaxil_a Q(s, a)$.

In the recent years, RL algorithms have been studied extensively in a branch referred to as Deep RL to deal with non-tabular state spaces with the aid of non-linear function approximation.
Deep Q-Networks (DQN) \cite{DBLP:journals/corr/MnihKSGAWR13} is a substantial one among these, that serves as a strong baseline in the domains with discrete actions.
In this off-policy algorithm, $Q(s, a)$ values are approximated with a deep neural network with weights $\theta$.
By using the transitions stored in a replay buffer, $\theta$ is learned by minimising the loss terms $(r_{k+1} + \gamma \max_{a'} Q_{\bar{\theta}}(s_{k+1}, a') - Q_{\theta}(s_{k}, a))^2$ with stochastic gradient descent, where $\bar{\theta}$ stands for the periodically updated copy of $\theta$.
This is the target network trick that DQN incorporates to battle the RL induced non-stationarity in the approximator learning.
Another critical component is the replay buffer, that lets the agent save samples to be learned from over a long course by also breaking the non-i.i.d. property of sequential collection.
DQN's success has led it to be studied and enhanced further over years.
The most prominent of these are summarised in Rainbow DQN \cite{DBLP:conf/aaai/HesselMHSODHPAS18}.
Among these, we employ Dueling Networks \cite{DBLP:conf/icml/WangSHHLF16} and Double Q-learning \cite{DBLP:conf/aaai/HasseltGS16} in this paper.

\subsection{Action Advising}

Action advising \cite{DBLP:conf/atal/TorreyT13} is a peer-to-peer knowledge exchange framework that requires only a common set of actions and a communication protocol between the peers.
According to this, a learning peer (student) receives advice in the form of actions from a more knowledgeable peer (teacher) to accelerate its learning.
These advice actions are generated directly from the teacher's decision-making policy; therefore, it is important for the student and the teacher to have the same goal in the task they are performing in.
A key property that distinguishes this approach from the similar frameworks such as Policy Reuse \cite{DBLP:conf/atal/FernandezV06} is the notion of budget constraint.
By considering the realistic scenarios where it is not possible to reliably exchange information, the number of interactions in this framework is also limited with a budget.
Consequently, the algorithms that operate in this problem setup should be capable of determining the most appropriate moments to exchange advice either from the perspective of teacher (teacher-initiated), student (student-initiated) or both (jointly-initiated).

\section{Proposed Algorithm} \label{sec:proposed_algorithm}

We adopt the MDP formalisation presented in Section~\ref{sec:background} in our problem definition.
The setup in this study includes an off-policy deep RL agent (student) with policy $\pi_S$ learning to perform some task $T$ in an environment with continuous state space and discrete actions.
There is also another agent with policy $\pi_T$ (teacher) that is competent in $T$.
The teacher is isolated from the environment itself, but is reachable by the student via a communication channel for a limited number of times defined by the advising budget $b$.
By using this mechanism, the student can request action advice $a_T = \pi_T(s)$ for its current state $s$.
The objective of the student in this problem is to maximise its learning performance in $T$ by timing these interactions to make the most efficient use of $\pi_T$.

Our approach provides a unified solution for addressing \emph{when to ask for advice} and \emph{how to leverage the advice} questions.
In addition to the RL algorithm, the student is equipped with a neural network $G_\omega$ with weights $\omega$ that are not shared with the RL model in any way.
The student also has a transitions buffer $D$ with no capacity limit that holds the collected state-advice pairs.
By using the samples in $D$, $G_\omega$ is trained periodically to provide the student an up-to-date imitation model of $\pi_T$ to make it possible to reuse the previously provided advice.
Moreover, $G_\omega$ is also used to determine what advice to collect by being regarded as a representation of $D$'s contents.
Obviously, making these decisions require $G_\omega$ to have a form of awareness of what it is trained on (in terms of samples).
Therefore, $G_\omega$ employs Dropout regularisation in the fully-connected layers to have an estimation of epistemic uncertainty denoted by $G_{\omega}^{\mu}(s)$ for any state $s$ as it is done in \cite{DBLP:conf/icml/GalG16}\cite{aawithimitation}.
None of these components share anything with or require access to the student's RL algorithm.
This is especially advantageous when it comes to pairing up our approach with different RL methods.

At the beginning of the student's learning process, $\omega$ is initialised randomly and $D = \emptyset$.
Then, at every timestep $t$ in state $s_t$, the student goes through $3$ stages of our algorithm: Collection, Imitation, Reuse.
The remainder of this section describes these stages with the line number references to the complete flow of our algorithm summarised in Algorithm~\ref{alg:proposed_approach}.

The collection stage (lines 13-19) remains active from the beginning until the student runs out of its advising budget $b$.
At this step, the student attempts to collect advice if its current state has not been advised before.
This is determined by the value of $G_{\omega}^{\mu} (s_t)$.
If it is higher than the uncertainty threshold $\tau$ (which is set automatically in the imitation stage), it is decided that $s_t$ is has not been advised before; thus, the student proceeds with requesting advice.
However, if $\tau$ is undetermined, this request is carried out without performing any uncertainty check.

The imitation module is responsible for training $G_\omega$ and tuning $\tau$ accordingly.
This stage (lines 20-22) is always active, but it is only triggered when these conditions that are checked at every timestep $t$ are met:
the student has collected $n_{min}$ new samples in $D$ (since the last imitation) or the student has taken $t_{min}$ steps (since the last imitation) with at least $n_{min} / 2$ new samples in $D$.
Here, $n_{min}$ and $t_{min}$ are hyperparameters.
These are set in order to keep the number of imitation processes within a reasonable number while also ensuring $G_\omega$ remains up-to-date with the collected advice.
On one hand, if $G_\omega$ was updated for every new state-advice pair, it would be a very accurate model of $D$'s contents, but the total training times would be a significant computational burden.
On the other hand, if $G_\omega$ was updated infrequently, it would not cause any computational setbacks; however, it would not be a good representation of the collected advice either.

Once the imitation is triggered, $G_\omega$ is trained for $k_{init}$ iterations (if it is the first ever training; else, for $k_{periodic}$ iterations) with the minibatches of samples drawn randomly from $D$.
This process resembles the simplest form of behavioural cloning where the supervised negative log-likelihood loss $\mathcal{L} (\omega) = \sum_{(s,a) \in D}^{} -log G_{\omega}(a \mid s)$ is minimised.
Afterwards, $\tau$ is updated automatically to be compatible with the new state of this imitation network.
This is done by measuring $G_{\omega}^{\mu}$ and storing them in a set $U$ for each $s_i$ in $D$ that satisfies $a_i = \argmaxil_a G_{\omega}(a \mid s_i)$.
Then, the uncertainty value that corresponds to the $p^{th}$ percentile (hyperparameter) in the ascending-order sorted $U$ is assigned to $\tau$.
We do this to pick a threshold $\tau$ such that $G_\omega$ can consider these samples it classifies correctly as ``known'' while leaving a small portion that are likely to be outliers out, when $G_{\omega}^{\mu}$ is compared with $\tau$.
This approach could be further developed by also considering the true-positive and false-positive rates, however, we opted for a simpler approach in this study.

Finally, the reuse stage (lines 23-26) handles the execution of the imitated advice whenever appropriate, to aid the student in efficient exploration.
It becomes active as soon as the imitation model $G_\omega$ is trained for the first time.
Then, whenever $G_{\omega}^{\mu}(s_t) < \tau$ (i.e. $G_{\omega}$ is familiar with $s_t$), no advice collection is occurred at $t$ and reusing is enabled for this particular episode, the student executes the imitated advice $\argmaxil_a G_{\omega}(a \mid s_t)$.
Unlike \cite{aawithimitation}, we do not limit advice reusing to the exploration stage of learning, e.g. the period $\epsilon$ is annealed to its final value in $\epsilon$-greedy.
Instead, we define a reuse schedule that is independent than the underlying RL algorithm's exploration strategy.
At the beginning of each episode, the agent either enables reuse module with a probability of $\rho$ (set as $\rho_{init}$ initially).
This value is decayed until it reaches its final value $\rho_{final}$ over $t_\rho$ steps, similarly to $\epsilon$-greedy annealing.
This approach further eliminates the dependency of our algorithm to the RL algorithm's exploration strategy.

\begin{algorithm}[!t]
\caption{Learning on a Budget via Teacher Imitation}
\label{alg:proposed_approach}
\begin{algorithmic}[1]
    \STATE {\bfseries Input:}  
    action advising budget $b$, 
    student policy $\pi_{S}$, 
    teacher policy $\pi_{T}$, 
    number of training iterations $t_{max}$,
    initial reuse probability $\rho_{init}$,
    final reuse probability $\rho_{final}$,
    $\rho$ decaying steps $t_\rho$,
    imitation network $G_{\omega}$,
    number of imitation training iterations $k_{init}$ and $k_{periodic}$,
    number of new samples and steps to trigger imitation $n_{min}$ and $t_{min}$

    \STATE $D \leftarrow \emptyset$ \algorithmiccomment{initialise empty state-advice buffer}
    \STATE \emph{reuse\_enabled} $\leftarrow False$ \algorithmiccomment{disable reuse by default}
    \STATE $\tau \leftarrow None$ \algorithmiccomment{no valid threshold}
    \STATE $\rho \leftarrow \rho_{init}$ \algorithmiccomment{set reuse probability with the initial value}
    \STATE $n_{last} \leftarrow 0$

    \FOR{training steps $t \in \{1, 2, \ldots t_{max}\}$}   
        \IF{$Env$ is reset}
            \STATE Set \emph{reuse\_enabled} $True$ with $\epsilon_{reuse}$ probability 
            \STATE Get observation $s_t \sim Env$ if $Env$ is reset
        \ENDIF
        
        \STATE $a_t \leftarrow None$ \algorithmiccomment{action is not determined yet}
        
        \dotfill
        
        \algorithmiccommentn{Collection}
        \IF{\emph{reuse\_enabled} is $True$ {\bf and} $b > 0$}
            \IF{$G_{\omega}$ is not trained {\bf or} $G_{\omega}^{\mu}(s_t) > \tau$}
                \STATE $a_t \leftarrow \pi_{T}(s_t)$ \algorithmiccomment{collect advice}
                \STATE Add $\langle s_t, a_{t}\rangle$ to $D$
                \STATE $b \leftarrow b - 1$ \algorithmiccomment{decrease the budget}
            \ENDIF
        \ENDIF
        
        \dotfill
      
        \algorithmiccommentn{Imitation}
        \IF{$\vert D \vert - n_{last} \geqslant n_{train}$ {\bf or}\\ 
        \quad ($\vert D \vert - n_{last} \geqslant n_{train} / 2$ {\bf and} $t - t_{train} \geqslant test $)} 
            \STATE Train $G_{\omega}$ with $D$ for $k_{init}$ or $k_{periodic}$ iterations
            \STATE $n_{last} \leftarrow \vert D \vert$
            \STATE Determine $\tau$ as described in Section~\ref{sec:proposed_algorithm}
        \ENDIF
        
        \dotfill
        
        \algorithmiccommentn{Reuse}
        \IF{\emph{reuse\_enabled} is $True$ {\bf and} $a_t$ is $None$ {\bf and}\\
        \quad $G_{\omega}$ is trained {\bf and} $G_{\omega}^{\mu}(s_t) < \tau$}
            \STATE $a_t \leftarrow \argmax{a} G_{\omega}(a \mid s_t)$
        \ENDIF
        \STATE Decay $\rho$ w.r.t. pre-defined schedule if $\rho > \rho_{final}$
        
        \dotfill
    
        \IF{$a_t$ is $None$}
                \STATE $a_t \leftarrow \pi_{S}(s_t)$ \algorithmiccomment{self policy}
        \ENDIF 
        
        \STATE Execute $a_t$ and obtain $r_t$, $s_{t+1} \sim Env$    
        \STATE Update the RL algorithm, e.g. DQN
        \STATE $s_t \leftarrow s_{t+1}$
    \ENDFOR
\end{algorithmic}
\end{algorithm}
\section{Experimental Setup} \label{sec:experiments}

We designed our experiments to answer the following questions about our proposal:
\begin{itemize}
\item How does our automatic threshold tuning perform against the manually-set ones in terms reuse accuracy and learning performance?
\item How much does using advice imitation model to drive the advice collection process help with collecting more diverse state-advice dataset?
\item Does collecting a dataset with more diverse samples make any significant impact on the learning performance?
\item How much does every particular modification contribute in the final performance?
\end{itemize}

In the remainder of this section, we first describe our evaluation domain. 
Then, we provide the details our experimental process along with the substantial implementation details\footnote{\text{The code for our experiments can be found at }\url{https://github.com/ercumentilhan/advice-imitation-reuse}}.

\subsection{Evaluation Domain}

\begin{figure}[t]
\centering
\begin{subfigure}[b]{0.19\linewidth} 
    \centering
    \setlength{\fboxsep}{0pt}\fbox{\includegraphics[width=0.85\textwidth]{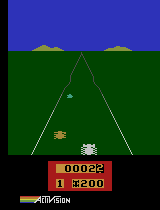}}
    \caption{Enduro}
\end{subfigure}
\begin{subfigure}[b]{0.19\linewidth} 
    \centering
    \setlength{\fboxsep}{0pt}\fbox{\includegraphics[width=0.85\textwidth]{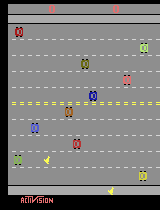}}
    \caption{Freeway}
\end{subfigure}
\begin{subfigure}[b]{0.19\linewidth}
    \centering
    \setlength{\fboxsep}{0pt}\fbox{\includegraphics[width=0.85\textwidth]{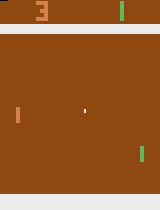}}
    \caption{Pong}
\end{subfigure}
\begin{subfigure}[b]{0.19\linewidth}
    \centering
    \setlength{\fboxsep}{0pt}\fbox{\includegraphics[width=0.85\textwidth]{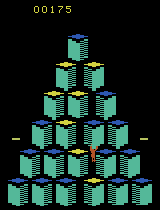}}
    \caption{Q*bert}
\end{subfigure}
\begin{subfigure}[b]{0.19\linewidth}
    \centering
    \setlength{\fboxsep}{0pt}\fbox{\includegraphics[width=0.85\textwidth]{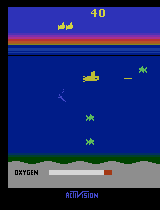}}
    \caption{Seaquest}
\end{subfigure}
\caption{Example RGB frames (observations) taken from the Arcade Learning Environment games Enduro (a), Freeway (b), Pong (c), Q*bert (d) and Seaquest (e).}
\label{fig_envs}
\end{figure}

In order to have the adequate amount of difficulty that is relevant to the modern deep RL algorithms, we chose the widely experimented Arcade Learning Environment (ALE) \cite{DBLP:journals/jair/BellemareNVB13} that contains more than $60$ Atari games as our testbed.
We picked $5$ well-known games among these, namely Enduro, Freeway, Pong, Q*bert, Seaquest, which involve different mechanics and present various learning challenges.
In each of these games, the agent receives observations as RGB images with a size of $160 \times 210 \times 3$.
These observations are converted into $80 \times 80 \times 1$ grayscale images to reduce the amount of representational complexity, and are also stacked as the most recent $4$ frames to eliminate the effect of partial observability.
Furthermore, since these games are originally processed at high frame-per-second rates with very little differences between two consecutive frames, agent actions are repeated for $4$ frames by skipping frames.
Consequently, the final $84 \times 84 \times 4$ sized observations the agent gets are built with the most recent $16$ game frames.
The range of rewards in these games are also different and unbounded.
Therefore, to facilitate the stability of learning, they are clipped to be in $[-1, 1]$ before they are provided to the agent.
The game episodes are limited to last for $108$k frames ($27$k agent steps) at maximum.

\subsection{Settings and Procedure}

We experiment with an extensive set of agents to be able to determine the most beneficial enhancements included in our algorithm.
The student agent variants we compare in our experiments are as follows:
\begin{itemize}
\item \textbf{No Advising (NA):} No form of action advising is employed, the agent relies on its RL algorithm only.
\item \textbf{Early Advising (EA):} The student asks for advice greedily until its budget runs out. There is no further utilisation of advice beyond their execution at the time of collection. This is a simple yet well-performing heuristic.
\item \textbf{Random Advising (RA):} The student asks for advice randomly with $0.5$ probability.
This heuristic uses the intuition that spacing out requests may yield more diverse and information rich advice. 
\item \textbf{{EA + Advice Reuse} (AR):} The agent employs the previously proposed advice reuse approach \cite{aawithimitation}. Advice is collected with early advising strategy, and the teacher is imitated with these advice. Then, advice are reused in place of the random exploration actions in approximately $0.5$ of the episodes.
\item \textbf{AR + Automatic Threshold Tuning ({AR+A}):} AR is combined with our automatic threshold tuning technique.
\item \textbf{AR+A + Extended Reuse (AR+A+E):} AR is combined with both our automatic threshold tuning technique and the extended reusing scheme.
\item \textbf{Advice Imitation \& Reuse (AIR):} This agent mode incorporates all of our proposed enhancements (as detailed in Section~\ref{sec:proposed_algorithm}). On top of AR+A+E, this mode also uses the imitation module's uncertainty to drive the advice collection process instead of relying on early advising.
\end{itemize}

We test the agents in learning sessions with length of $5$M steps (equals to $20$M game frames due to frame skipping) with an advising budget of $25$k that corresponds to only $0.5\%$ of the total number of steps in a session.
At every $50$k$^{th}$ step, the agents are evaluated in a separate set of $10$ episodes by having their action advising and exploration mechanisms disabled.
The cumulative rewards obtained in these episodes are averaged and recorded as evaluation scores for that corresponding learning session step.
This lets us measure the actual learning progress of the agents as the main performance metric.

The deep RL algorithm of the student agents is Double DQN with a neural network structure comprised of $3$ convolutional layers ($32$ $8 \times 8$ filters with a stride of $4$ followed by $64$ $4 \times 4$ filters with a stride of $2$ followed by $64$ $3 \times 3$ filters with a stride of $1$) and fully-connected layers with a single hidden layer ($512$ units) and dueling stream output.
For exploration, $\epsilon$-greedy strategy with linearly decaying $\epsilon$ is adopted.
The teacher agents are generated separately for each of the games prior to the experiments, by using the identical DQN algorithm and structure with the student.
Even though the resulting agents are not necessarily at super-human levels achievable by DQN, they have competent policies that can achieve the evaluation scores of $1556$, $28.8$, $12$, $3705$, $8178$ for Enduro, Freeway, Pong, Q*bert, Seaquest, respectively.

As we described in Section~\ref{sec:proposed_algorithm}, our approach requires the student to be equipped with an additional behavioural cloning module that includes a neural network.
We used the identical neural network structure to the student's DQN model except for the dueling streams.
Fully-connected layers of this network are enhanced with Dropout regularisation with a dropout rate of $0.35$ and the number of forward passes to measure the epistemic uncertainty via variance is set at $100$.

The uncertainty threshold for AR is set as $0.01$ for every game.
Determining a reasonable value for this parameter requires to access the tasks briefly, which we have performed prior to the experiments; even though this will not be reflected at the numerical results, it should be noted that this is a critical disadvantage of AR.
The automatic threshold tuning percentile used in AR+A, AR+A+E, AIR is set as $90$.
This is a very straightforward hyperparameter to adjust compared to the (manual) uncertainty threshold itself and can potentially be valid in a wide variety of tasks.
For the extensive reuse scheme in AR+A+E and AIR, we set $\rho_{init}$ and $\rho_{final}$ as $0.5$ and $0.1$, respectively.
We defined the annealing schedule to begin at $500$k$^{th}$ step and last until $2$M$^{th}$ step.
For the imitation triggering conditions in AIR, $n_{min}$ is set as $2.5$k (samples) and $t_{min}$ is set as $50$k (timesteps).
Finally, the number of imitation network training iterations is set as $200$k for the initial one (applies to all modes but NA, EA and RA) and $50$k for the periodic ones (only applies to AIR).

All of the aforementioned hyperparameters reported in this section are set empirically prior to the experiments and are kept the same across every game.
The most significant ones among the unmentioned hyperparameters of the student's learning components are presented in Table~\ref{table:hyperparameters}.

\begin{table}[!t]
	\centering
	\caption{Hyperparameters of the student's DQN and imitation module (for AR, AR+A, AR+A+E, AIR).}
	\label{table:hyperparameters}
	\begin{tabular}{l|c}  	
		Hyperparameter name & Value \\
  	    \cmidrule(r){1-2}
  	    Discount factor $\gamma$ & $0.99$ \\
  	    Learning rate & $625 \times 10^{-7}$ \\
  	    Minibatch size & $32$ \\
  	    Replay memory min. size and capacity & $50$k, $500$k \\
		Target network update period & $7500$ \\
		$\epsilon$ initial, $\epsilon$ final, $\epsilon$ decay steps & $1.0$, $0.01$, $500$k \\
		\cmidrule(r){1-2}
		Learning rate & $0.0001$ \\
		Minibatch size & $32$ \\
	\end{tabular}
\end{table}

\section{Results} \label{sec:results}

\begin{figure*}[!t]
\centering
\includegraphics[width=1.0\textwidth]{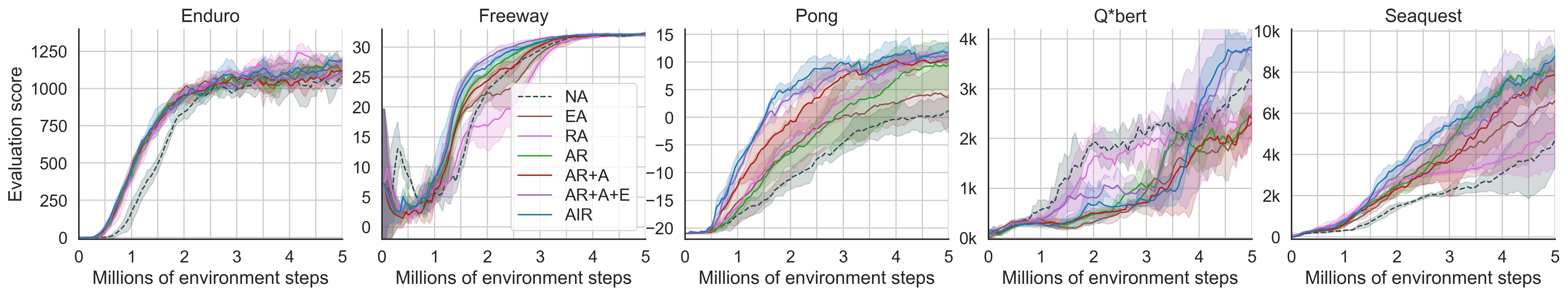}
\caption{
Evaluation scores obtained by the agent modes NA, EA, RA, AR, AR+A, AR+A+E, AIR in the ALE games Enduro, Freeway, Pong, Q*bert, Seaquest.
Shaded areas show the standard deviation across $3$ runs.
}
\label{fig:results_eval}
\end{figure*}

\begin{figure*}[!t]
\centering
\includegraphics[width=1.0\textwidth]{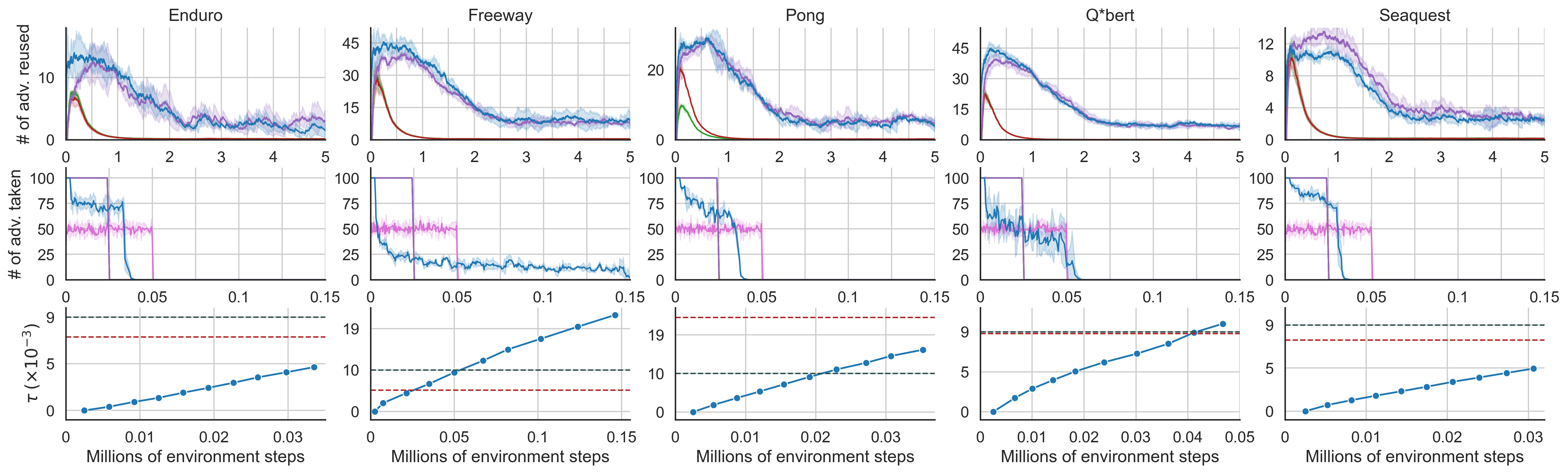}
\caption{
Top and middle rows show the plots for the number of advice reuses and advice collections (for the first $5$M and $150$k steps, respectively) in every $100$ steps performed by the relevant student modes.
The legend for colours is the same as Figure~\ref{fig:results_eval} and the shaded areas show the standard deviation across $3$ different runs.
Bottom row shows the values of $\tau$ across a single learning session, with different timestep scopes determined by the length of AIR's advice collection stage.
Blue colour represents AIR, the dashed grey lines represent AR, and the dashed red lines represent AR+A and AR+A+E.
}
\label{fig:results_reuse}
\end{figure*}

\begin{figure}[t]
\centering
\begin{subfigure}[b]{0.48\linewidth} 
    \centering
    \setlength{\fboxsep}{0pt}\fbox{\includegraphics[width=0.95\textwidth]{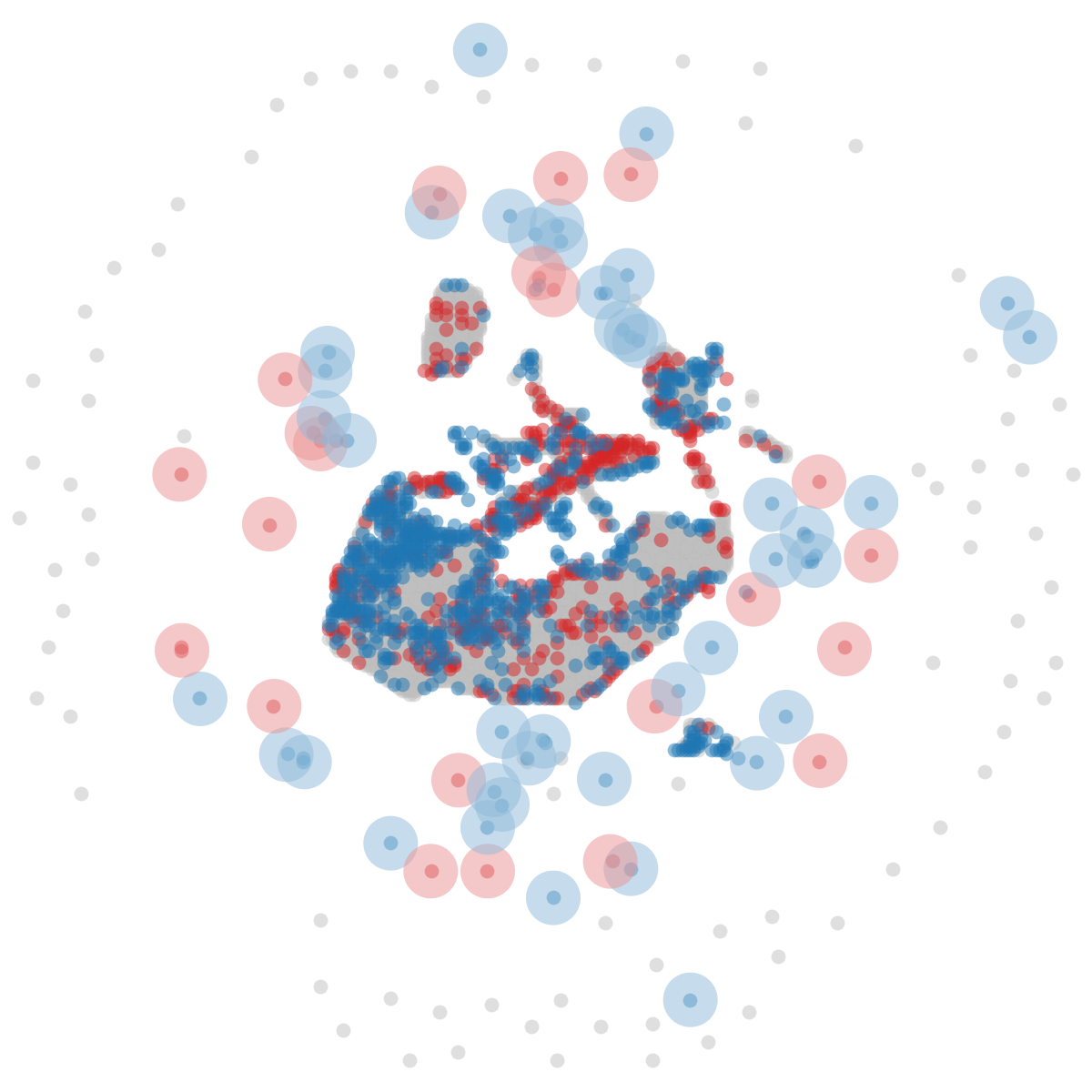}}
    \caption{AIR vs. EA}
\end{subfigure}
\begin{subfigure}[b]{0.48\linewidth} 
    \centering
    \setlength{\fboxsep}{0pt}\fbox{\includegraphics[width=0.95\textwidth]{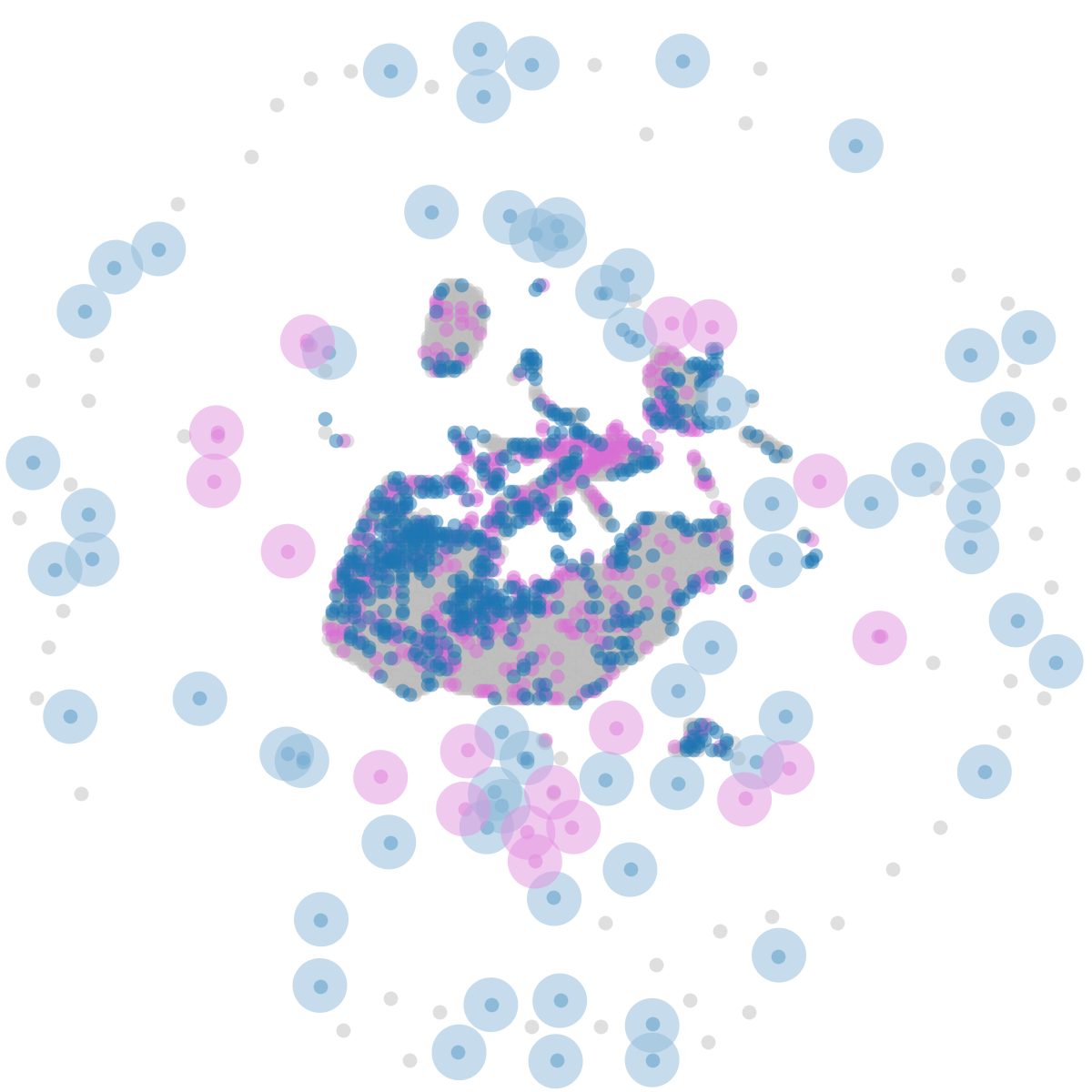}}
    \caption{AIR vs. RA}
\end{subfigure}
\caption{UMAP \cite{DBLP:journals/corr/abs-1802-03426} embeddings of the advice collected states in Seaquest by AIR (blue) vs. EA (red) in (a) and AIR (blue) vs. RA (pink) in (b), where the samples in common are shown in grey. Areas denoted with larger circles are the outliers covered only by either AIR (blue), EA (red) or RA (pink).}
\label{fig:umap}
\end{figure}

\begin{table}[t]
	\centering
	\caption{
	Final evaluation scores, percentages of advice reuses in the total environment steps, percentages of advice reuse accuracies achieved by NA, EA, RA, AR, AR+A, AR+A+E, AIR (+ signs are omitted) in $5$ ALE games aggregated over $3$ runs.
	The standard deviations across runs are indicated with $\pm$. The best scores/accuracies are denoted in bold.
	}
	\label{table:results}
	\begin{tabular}{p{0.78cm}p{0.6cm}cp{1.45cm}p{1.5cm}}  
	\toprule
	& & \multicolumn{1}{c}{Evaluation Score} & \multicolumn{2}{c}{Advice Reuse}\\
	\cmidrule(r){3-3}
	\cmidrule(r){4-5}
	\multicolumn{1}{c}{{Game}} & Mode & \multicolumn{1}{c}{Final} 
	& \multicolumn{1}{c}{Ratio ($\%$)} & \multicolumn{1}{c}{Accuracy (\%)}\\	
	\midrule
	
	\multirow{7}{*}{Enduro} 
    & NA
    & $1066.34 \pm 37.3$
    & \multicolumn{1}{c}{{---}}
    & \multicolumn{1}{c}{{---}}
    \\
    & EA
    & $1131.13 \pm 69.0$
    & \multicolumn{1}{c}{{---}}
    & \multicolumn{1}{c}{{---}}
    \\
    & RA
    & $1170.60 \pm 19.0$
    & \multicolumn{1}{c}{{---}}
    & \multicolumn{1}{c}{{---}}
    \\
    & AR
    & $1127.72 \pm 26.9$
    & $0.49 \pm 0.03$
    & $70.06 \pm 0.6$
    \\
    & ARA
    & $1117.93 \pm 59.0$
    & $0.45 \pm 0.04$
    & $72.36 \pm 0.3$
    \\
    & ARAE
    & $1102.44 \pm 62.4$
    & $4.97 \pm 0.44$
    & $\bm{75.24 \pm 0.5}$
    \\
    & AIR
    & $\bm{1184.02 \pm 19.6}$
    & $4.79 \pm 0.44$
    & $73.63 \pm 0.2$
    \\
 
    \cmidrule(r){1-5}

	\multirow{7}{*}{Freeway}
    & NA
    & $31.98 \pm 0.1$
    & \multicolumn{1}{c}{{---}}
    & \multicolumn{1}{c}{{---}}
    \\
    & EA
    & $32.09 \pm 0.1$
    & \multicolumn{1}{c}{{---}}
    & \multicolumn{1}{c}{{---}}
    \\
    & RA
    & $32.02 \pm 0.2$
    & \multicolumn{1}{c}{{---}}
    & \multicolumn{1}{c}{{---}}
    \\
    & AR
    & $32.06 \pm 0.1$
    & $1.60 \pm 0.11$
    & $93.02 \pm 0.3$
    \\
    & ARA
    & $\bm{32.26 \pm 0.1}$
    & $1.53 \pm 0.10$
    & $93.48 \pm 0.1$
    \\
    & ARAE
    & $32.14 \pm 0.0$
    & $15.76 \pm 0.08$
    & $\bm{95.43 \pm 0.1}$
    \\
    & AIR
    & $32.14 \pm 0.1$
    & $17.89 \pm 0.47$
    & $95.39 \pm 0.2$
    \\

    \cmidrule(r){1-5} 
    
	\multirow{7}{*}{Pong}
	& NA
    & $0.95 \pm 2.4$
    & \multicolumn{1}{c}{{---}}
    & \multicolumn{1}{c}{{---}}
    \\
    & EA
    & $3.73 \pm 4.9$
    & \multicolumn{1}{c}{{---}}
    & \multicolumn{1}{c}{{---}}
    \\
    & RA
    & $11.48 \pm 0.2$
    & \multicolumn{1}{c}{{---}}
    & \multicolumn{1}{c}{{---}}
    \\
    & AR
    & $9.41 \pm 3.5$
    & $0.52 \pm 0.02$
    & $79.80 \pm 0.5$
    \\
    & ARA
    & $10.48 \pm 0.4$
    & $0.93 \pm 0.01$
    & $75.07 \pm 0.7$
    \\
    & ARAE
    & $11.21 \pm 1.2$
    & $10.4 \pm 0.59$
    & $79.00 \pm 0.1$
    \\
    & AIR
    & $\bm{11.81 \pm 1.2}$
    & $9.98 \pm 0.36$
    & $\bm{80.46 \pm 0.8}$
    \\

    \cmidrule(r){1-5} 
    
	\multirow{7}{*}{Q*bert}
    & NA
    & $3154.91 \pm 408.9$
    & \multicolumn{1}{c}{{---}}
    & \multicolumn{1}{c}{{---}}
    \\
    & EA
    & $2277.98 \pm 300.5$
    & \multicolumn{1}{c}{{---}}
    & \multicolumn{1}{c}{{---}}
    \\
    & RA
    & $2528.70 \pm 505.4$
    & \multicolumn{1}{c}{{---}}
    & \multicolumn{1}{c}{{---}}
    \\
    & AR
    & $2434.70 \pm 54.8$
    & $0.93 \pm 0.04$
    & $80.37 \pm 0.6$
    \\
    & ARA
    & $2359.39 \pm 371.9$
    & $0.93 \pm 0.02$
    & $78.86 \pm 0.7$
    \\
    & ARAE
    & $3763.72 \pm 340.3$
    & $14.57 \pm 0.25$
    & $\bm{92.87 \pm 0.7}$
    \\
    & AIR
    & $\bm{3814.34 \pm 134.6}$
    & $15.16 \pm 0.21$
    & $92.83 \pm 0.2$
    \\
    
    \cmidrule(r){1-5} 
    
	\multirow{7}{*}{Seaquest}
    & NA
    & $4496.41 \pm 1101.0$
    & \multicolumn{1}{c}{{---}}
    & \multicolumn{1}{c}{{---}}
    \\
    & EA
    & $6538.18 \pm 1445.1$
    & \multicolumn{1}{c}{{---}}
    & \multicolumn{1}{c}{{---}}
    \\
    & RA
    & $5033.04 \pm 1413.3$
    & \multicolumn{1}{c}{{---}}
    & \multicolumn{1}{c}{{---}}
    \\
    & AR
    & $8053.93 \pm 935.9$
    & $0.61 \pm 0.03$
    & $72.96 \pm 0.8$
    \\
    & ARA
    & $7851.03 \pm 556.7$
    & $0.56 \pm 0.03$
    & $\bm{76.29 \pm 0.7}$
    \\
    & ARAE
    & $8082.69 \pm 1105.2$
    & $5.93 \pm 0.49$
    & $74.18 \pm 1.5$
    \\
    & AIR
    & $\bm{8614.04 \pm 268.7}$
    & $4.79 \pm 0.20$
    & $74.87 \pm 0.5$
    \\

	\bottomrule
	\end{tabular}
\end{table}

The results of our experiments in Enduro, Freeway, Pong, Q*bert and Seaquest are presented in several plots to let us analyse the performance of the student modes NA, EA, RA, AR, AR+A, AR+A+E, AIR in different aspects.
Figure~\ref{fig:results_eval} contains the evaluation scores plots.
In Figure~\ref{fig:results_reuse}, plots for the number of advice reuses per $100$ steps (top row) performed by AR, AR+A, AR+A+E, AIR; plots for the number of advice collections per $100$ steps performed by NA, RA, AR, AR+A, AR+A+E, AIR (middle row); plots for the values of the $\tau$ hyperparameter (uncertainty threshold) used by AR, AR+A, AR+A+E, AIR (bottom row) in a single run are shown.
The shaded areas in these plots show the standard deviation across $3$ runs.
The final evaluation scores, percentage of advice reuses in total number of environment steps as well as their accuracies are presented in Table~\ref{table:results}.
Finally, in order to highlight the differences in the advice-collected state diversity of EA (identical collection strategy to AR, AR+A, AR+A+E), RA and AIR, these states from a single Pong run are visualised with the aid of UMAP \cite{DBLP:journals/corr/abs-1802-03426} dimensionality reduction technique in Figure~\ref{fig:umap}.
Here, the scatter plot on the left compares AIR (blue) vs. EA (red) and the one on the right compares AIR (blue) vs. RA (pink); and the samples collected in common are shown in grey.

We first analyse the learning performances via evaluation scores.
In Enduro, all methods but NA have a very similar learning speed and final scores, with AIR being slightly ahead of the rest.
In Freeway, they all achieve nearly the same final scores, but they are distinguishable with small differences in learning speed where AR+A+E is on the top followed by other advanced student modes AIR, AR, AR+A.
When we move to Pong, Q*bert and Seaquest, we finally see the student mode performances to be more distinctive.
Even though the basic heuristics (RA and EA) show that a little number of advice from a competent policy can make substantial boosts on learning, these modes fall behind of those that employ advice reuse and fail to be a reliable choice, i.e. performing worse than NA in Freeway and Q*bert.
Overall, the best mode AIR and the runner-up AR+A+E are ahead of all, with AR and AR+A following them.

Among the advice reuse approaches, we see that the most beneficial modification is the extended reuse schedule (+E) as it is highlighted by the difference between AR+A and AR+A+E.
Defining such a schedule independently from the student's RL exploration strategy involves using some extra hyperparameters, nevertheless, they are rather trivial to set arbitrarily.
The trends in the advice reuse plots (Figure~\ref{fig:results_reuse}, top row) show how these schedules differ.
The versions with +E (AIR and AR+A+E) yield around $10\times$ more reusing, which apparently plays an important role in the performance improvement.
However, it is still not clear how to define the optimal reuse schedules.

Automatic threshold tuning (+A) also performs comparably, if not better, with the manual tuning approach (AR) as we can observe in the evaluation scores.
Additionally, AR+A managed to achieve very similar reuse accuracies with AR; this also supports its success.
When we examine the $\tau$ values, we see AR+A determined values that are close to the hand-tuned ones, except for the case in Pong where the difference is more significant.
This reflected in reuse trend and evaluation performance, giving AR+A a very advantageous head start.
These results support the idea that +A a far more preferable approach considering how problematic it can be to tune the sensitive $\tau$ threshold manually.
For instance, if they were to deployed in some significantly different domains as they are, then we could potentially see AR+A coming far ahead of AR with a poorly tuned $\tau$. 
Furthermore, the periodic imitation model updates incorporated in AIR makes +A an essential component.
In the bottom row of Figure~\ref{fig:results_reuse}, we also show how AIR changes its $\tau$ values over time as it collects more advice samples and updates its imitation model accordingly.
Clearly, it is very difficult to manage these changes manually.

Finally, we also see that collecting advice by utilising the imitation model's uncertainty (as it is done by AIR) contributes to the agent's learning.
When we look at the advice collection plots, we see $3$ different types of behaviours: early collection (EA, AR, AR+A, AR+A+E), random collection (RA), and AIR.
Even though they seem to be occurring mostly in the same time windows, AIR does this in an uncertainty-aware fashion; hence the decreasing collection rate over time.
Freeway is the case in which AIR is very selective.
This is possibly due to the fact that in Freeway, the agent can traverse only a limited space which consequently reduces the diversity of the acquired observations.
Nonetheless, this is not reflected in the evaluation scores as dramatically due to this game being rather trivial to solve.
Another interesting observation is made by analysing the advice collected states in Seaquest by AIR, EA (which is identical to AR, AR+A, AR+A+E in terms of collection strategy), RA in a reduced dimensionality as seen in Figure~\ref{fig:umap}.
We chose Seaquest since it is the game where AIR is significantly ahead of AR+A+E, which can be credited to AIR's only difference from it (collection strategy).
We also include RA here mainly because it can potentially do better in acquiring different samples than EA. 
Here, the large circles denote the outliers (diverse samples) that are only covered only by either AIR, EA or RA.
These are the important bits to pay attention to and compare.
As it can be seen, AIR yields larger coverage, i.e. more diverse dataset of advice, in both cases against EA and RA collection strategies.
\section{Conclusions and Future Work} \label{sec:conclusions}

In this study, we proposed an automatic threshold tuning technique, an extended advice reusing schedule and an imitation model uncertainty-based advice collection procedure by extending the previously proposed advice reusing algorithm.
We also developed a combined approach by incorporating these components, that is able to collect a diverse set of advice to build a more widely applicable advice imitation model for advice reuse.

The experiments in $5$ different Atari games from the ALE domain have shown that our enhancements provide significant improvements over the baseline advice reuse method as well as the basic action advising heuristics.
First, being able to tune the uncertainty thresholds on-the-fly was observed to yield the learning performance of the carefully tuned threshold, which require unrealistic access to the tasks and extra effort to be adjusted.
Secondly, we found that having the advice reusing process span across a larger portion of the learning session rather than just the steps that involve random exploration can yield superior performance.
However, defining the best schedule for the maximum advice utilisation efficiency remains to be an open question.
Thirdly, the uncertainty-driven advice collection method was found to be successful way to improve the imitation module's dataset diversity.
Nevertheless, periodic training process can be improved with better incremental learning techniques to make a better use of this simultaneous collection-imitation idea.
Finally, our unified algorithm demonstrated state-of-the-art performance across $5$ Atari games by performing either on-par or better than its closest competitors.

The future extensions of this work can involve experimenting with more principled Policy Reuse approaches in the literature to further improve the advice reuse strategy.
Furthermore, it will be a worthwhile study to make the teacher imitation better at learning online from the new samples it acquires.
Finally, even though it is in the core motivation of our approach not to access and modify the agent's RL components, it will be beneficial to investigate Learning from Demonstrations techniques and their possible contributions in our framework.

\section*{Acknowledgment}
This research utilised Queen Mary's Apocrita HPC facility, supported by QMUL Research-IT. http://doi.org/10.5281/zenodo.438045

\printbibliography

@article{DBLP:journals/corr/abs-1812-02632,
  author    = {Si{-}An Chen and
               Voot Tangkaratt and
               Hsuan{-}Tien Lin and
               Masashi Sugiyama},
  title     = {Active Deep Q-learning with Demonstration},
  journal   = {CoRR},
  volume    = {abs/1812.02632},
  year      = {2018}
}

@article{DBLP:journals/corr/abs-1908-02388,
  author    = {Adrien Ali Ta{\"{\i}}ga and
               William Fedus and
               Marlos C. Machado and
               Aaron C. Courville and
               Marc G. Bellemare},
  title     = {Benchmarking Bonus-Based Exploration Methods on the Arcade Learning
               Environment},
  journal   = {CoRR},
  volume    = {abs/1908.02388},
  year      = {2019}
}

@inproceedings{DBLP:conf/aaai/HesterVPLSPHQSO18,
  author    = {Todd Hester and
               Matej Vecer{\'{\i}}k and
               Olivier Pietquin and
               Marc Lanctot and
               Tom Schaul and
               Bilal Piot and
               Dan Horgan and
               John Quan and
               Andrew Sendonaris and
               Ian Osband and
               Gabriel Dulac{-}Arnold and
               John P. Agapiou and
               Joel Z. Leibo and
               Audrunas Gruslys},
  title     = {Deep Q-learning From Demonstrations},
  booktitle = {Proceedings of the Thirty-Second {AAAI} Conference on Artificial Intelligence,
               (AAAI-18)},
  pages     = {3223--3230},
  publisher = {{AAAI} Press},
  year      = {2018}
}

@inproceedings{DBLP:conf/aaai/HasseltGS16,
  author    = {Hado van Hasselt and
               Arthur Guez and
               David Silver},
  title     = {Deep Reinforcement Learning with Double Q-Learning},
  booktitle = {Proceedings of the Thirtieth {AAAI} Conference on Artificial Intelligence},
  pages     = {2094--2100},
  publisher = {{AAAI} Press},
  year      = {2016}
 }

@inproceedings{DBLP:conf/icml/WangSHHLF16,
  author    = {Ziyu Wang and
               Tom Schaul and
               Matteo Hessel and
               Hado van Hasselt and
               Marc Lanctot and
               Nando de Freitas},
  title     = {Dueling Network Architectures for Deep Reinforcement Learning},
  booktitle = {Proceedings of the 33nd International Conference on Machine Learning,
               {ICML} 2016},
  series    = {{JMLR} Workshop and Conference Proceedings},
  volume    = {48},
  pages     = {1995--2003},
  year      = {2016}
}

@article{DBLP:journals/corr/abs-1810-12894,
  author    = {Yuri Burda and
               Harrison Edwards and
               Amos J. Storkey and
               Oleg Klimov},
  title     = {Exploration by Random Network Distillation},
  journal   = {CoRR},
  volume    = {abs/1810.12894},
  year      = {2018}
}

@inproceedings{DBLP:conf/ijcai/AmirKKG16,
  author    = {Ofra Amir and
               Ece Kamar and
               Andrey Kolobov and
               Barbara J. Grosz},
  title     = {Interactive Teaching Strategies for Agent Training},
  booktitle = {Proceedings of the Twenty-Fifth International Joint Conference on
               Artificial Intelligence, {IJCAI} 2016},
  pages     = {804--811},
  publisher = {{IJCAI/AAAI} Press},
  year      = {2016}
}

@inproceedings{DBLP:conf/atal/ZhuCLH20,
  author    = {Changxi Zhu and
               Yi Cai and
               Ho{-}fung Leung and
               Shuyue Hu},
  title     = {Learning by Reusing Previous Advice in Teacher-Student Paradigm},
  booktitle = {Proceedings of the 19th International Conference on Autonomous Agents
               and Multiagent Systems, {AAMAS} '20},
  pages     = {1674--1682},
  publisher = {International Foundation for Autonomous Agents and Multiagent Systems},
  year      = {2020}
}

@inproceedings{DBLP:conf/nips/Schaal96,
  author    = {Stefan Schaal},
  title     = {Learning from Demonstration},
  booktitle = {Advances in Neural Information Processing Systems 9, NIPS, Denver,
               CO, USA, December 2-5, 1996},
  pages     = {1040--1046},
  publisher = {{MIT} Press},
  year      = {1996}
}

@article{DBLP:journals/corr/MnihKSGAWR13,
  author    = {Volodymyr Mnih and
               Koray Kavukcuoglu and
               David Silver and
               Alex Graves and
               Ioannis Antonoglou and
               Daan Wierstra and
               Martin A. Riedmiller},
  title     = {Playing Atari with Deep Reinforcement Learning},
  journal   = {CoRR},
  volume    = {abs/1312.5602},
  year      = {2013}
}

@article{DBLP:journals/connection/TaylorCFVT14,
  author    = {Matthew E. Taylor and
               Nicholas Carboni and
               Anestis Fachantidis and
               Ioannis P. Vlahavas and
               Lisa Torrey},
  title     = {Reinforcement learning agents providing advice in complex video games},
  journal   = {Connect. Sci.},
  volume    = {26},
  number    = {1},
  pages     = {45--63},
  year      = {2014}
}

@book{sutton2018reinforcement,
  title={Reinforcement learning: An introduction},
  author={Sutton, Richard S and Barto, Andrew G},
  year={2018},
  publisher={MIT press}
}

@inproceedings{DBLP:conf/atal/SilvaGC17,
  author    = {Felipe Leno da Silva and
               Ruben Glatt and
               Anna Helena Reali Costa},
  title     = {Simultaneously Learning and Advising in Multiagent Reinforcement Learning},
  booktitle = {Proceedings of the 16th Conference on Autonomous Agents and {Multi-Agent}
               Systems, {AAMAS} 2017},
  pages     = {1100--1108},
  publisher = {{ACM}},
  year      = {2017}
}

@misc{ilhan2020studentinitiated,
      title={Student-Initiated Action Advising via Advice Novelty}, 
      author={Ercument Ilhan and Jeremy Gow and Diego Perez-Liebana},
      year={2021},
      eprint={2010.00381},
      archivePrefix={arXiv}
}

@inproceedings{zimmer2014teacher,
  title={{Teacher-Student Framework: A Reinforcement Learning Approach}},
  author={Zimmer, Matthieu and Viappiani, Paolo and Weng, Paul},
  booktitle={{AAMAS} Workshop Autonomous Robots and Multirobot Systems},
  year={2014}
}

@inproceedings{DBLP:conf/atal/TorreyT13,
  author    = {Lisa Torrey and
               Matthew E. Taylor},
  title     = {Teaching on a budget: agents advising agents in reinforcement learning},
  booktitle = {International conference on Autonomous Agents and Multi-Agent Systems,
               {AAMAS} '13, Saint Paul, MN, USA, May 6-10, 2013},
  pages     = {1053--1060},
  year      = {2013}
}

@inproceedings{DBLP:conf/cig/IlhanGP19,
  author    = {Erc{\"{u}}ment Ilhan and
               Jeremy Gow and
               Diego P{\'{e}}rez{-}Li{\'{e}}bana},
  title     = {Teaching on a Budget in Multi-Agent Deep Reinforcement Learning},
  booktitle = {{IEEE} Conference on Games, CoG 2019, London, United Kingdom, August
               20-23, 2019},
  pages     = {1--8},
  year      = {2019}
}

@article{DBLP:journals/jair/BellemareNVB13,
  author    = {Marc G. Bellemare and
               Yavar Naddaf and
               Joel Veness and
               Michael Bowling},
  title     = {The Arcade Learning Environment: An Evaluation Platform for General
               Agents},
  journal   = {J. Artif. Intell. Res.},
  volume    = {47},
  pages     = {253--279},
  year      = {2013}
}

@inproceedings{DBLP:conf/aaai/SilvaHKT20a,
  author    = {Felipe Leno da Silva and
               Pablo Hernandez{-}Leal and
               Bilal Kartal and
               Matthew E. Taylor},
  title     = {Uncertainty-Aware Action Advising for Deep Reinforcement Learning
               Agents},
  booktitle = {The Thirty-Fourth {AAAI} Conference on Artificial Intelligence, {AAAI}
               2020},
  pages     = {5792--5799},
  publisher = {{AAAI} Press},
  year      = {2020}
}

@article{DBLP:journals/neco/Pomerleau91,
  author    = {Dean Pomerleau},
  title     = {Efficient Training of Artificial Neural Networks for Autonomous Navigation},
  journal   = {Neural Com.},
  volume    = {3},
  number    = {1},
  pages     = {88--97},
  year      = {1991}
}

@inproceedings{DBLP:conf/icml/GalG16,
  author    = {Yarin Gal and
               Zoubin Ghahramani},
  title     = {Dropout as a Bayesian Approximation: Representing Model Uncertainty
               in Deep Learning},
  booktitle = {Proceedings of the 33nd International Conference on Machine Learning,
               {ICML} 2016, New York City, NY, USA, June 19-24, 2016},
  volume    = {48},
  pages     = {1050--1059},
  year      = {2016}
}

@article{DBLP:journals/corr/abs-1911-08265,
  author    = {Julian Schrittwieser and
               Ioannis Antonoglou and
               Thomas Hubert and
               Karen Simonyan and
               Laurent Sifre and
               Simon Schmitt and
               Arthur Guez and
               Edward Lockhart and
               Demis Hassabis and
               Thore Graepel and
               Timothy P. Lillicrap and
               David Silver},
  title     = {Mastering Atari, Go, Chess and Shogi by Planning with a Learned Model},
  journal   = {CoRR},
  volume    = {abs/1911.08265},
  year      = {2019}
}

@article{DBLP:journals/corr/abs-1910-07113,
  author    = {OpenAI and
               Ilge Akkaya and
               Marcin Andrychowicz and
               Maciek Chociej and
               Mateusz Litwin and
               Bob McGrew and
               Arthur Petron and
               Alex Paino and
               Matthias Plappert and
               Glenn Powell and
               Raphael Ribas and
               Jonas Schneider and
               Nikolas Tezak and
               Jerry Tworek and
               Peter Welinder and
               Lilian Weng and
               Qiming Yuan and
               Wojciech Zaremba and
               Lei Zhang},
  title     = {Solving Rubik's Cube with a Robot Hand},
  journal   = {CoRR},
  volume    = {abs/1910.07113},
  year      = {2019}
}

@inproceedings{aawithimitation,
  author    = {Erc{\"{u}}ment Ilhan and
               Jeremy Gow and
               Diego Perez-Liebana},
  title     = {Action Advising with Advice Imitation in Deep Reinforcement Learning},
  booktitle = {Proceedings of the 20th Conference on Autonomous Agents and {Multi-Agent}
               Systems, {AAMAS} 2021, May 3-7, 2021},
  %pages     = {1100--1108},
  publisher = {{IFAAMAS}},
  year      = {2021},
  %note      = {(To appear)}
}

@article{DBLP:journals/corr/abs-1802-03426,
  author    = {Leland McInnes and
               John Healy},
  title     = {{UMAP:} Uniform Manifold Approximation and Projection for Dimension
               Reduction},
  journal   = {CoRR},
  volume    = {abs/1802.03426},
  year      = {2018}
}

@inproceedings{DBLP:conf/atal/FernandezV06,
  author    = {Fernando Fern{\'{a}}ndez and
               Manuela M. Veloso},
  title     = {Probabilistic policy reuse in a reinforcement learning agent},
  booktitle = {5th International Joint Conference on Autonomous Agents and Multiagent
               Systems {(AAMAS} 2006), Hakodate, Japan, May 8-12, 2006},
  pages     = {720--727},
  publisher = {{ACM}},
  year      = {2006}
}

@inproceedings{DBLP:conf/aaai/HesselMHSODHPAS18,
  author    = {Matteo Hessel and
               Joseph Modayil and
               Hado van Hasselt and
               Tom Schaul and
               Georg Ostrovski and
               Will Dabney and
               Dan Horgan and
               Bilal Piot and
               Mohammad Gheshlaghi Azar and
               David Silver},
  title     = {Rainbow: Combining Improvements in Deep Reinforcement Learning},
  booktitle = {Proceedings of the Thirty-Second {AAAI} Conference on Artificial Intelligence,
               (AAAI-18)},
  pages     = {3215--3222},
  publisher = {{AAAI} Press},
  year      = {2018}
}

@article{DBLP:journals/nature/VinyalsBCMDCCPE19,
  author    = {Oriol Vinyals and
               Igor Babuschkin and
               Wojciech M. Czarnecki and
               Micha{\"{e}}l Mathieu and
               Andrew Dudzik and
               Junyoung Chung and
               David H. Choi and
               Richard Powell and
               Timo Ewalds and
               Petko Georgiev and
               Junhyuk Oh and
               Dan Horgan and
               Manuel Kroiss and
               Ivo Danihelka and
               Aja Huang and
               Laurent Sifre and
               Trevor Cai and
               John P. Agapiou and
               Max Jaderberg and
               Alexander Sasha Vezhnevets and
               R{\'{e}}mi Leblond and
               Tobias Pohlen and
               Valentin Dalibard and
               David Budden and
               Yury Sulsky and
               James Molloy and
               Tom L. Paine and
               {\c{C}}aglar G{\"{u}}l{\c{c}}ehre and
               Ziyu Wang and
               Tobias Pfaff and
               Yuhuai Wu and
               Roman Ring and
               Dani Yogatama and
               Dario W{\"{u}}nsch and
               Katrina McKinney and
               Oliver Smith and
               Tom Schaul and
               Timothy P. Lillicrap and
               Koray Kavukcuoglu and
               Demis Hassabis and
               Chris Apps and
               David Silver},
  title     = {Grandmaster level in {StarCraft II} using multi-agent reinforcement
               learning},
  journal   = {Nature},
  volume    = {575},
  number    = {7782},
  pages     = {350--354},
  year      = {2019}
}

\end{document}